\documentclass[letterpaper]{article}
\usepackage{aaai2027}
\usepackage[hyphens]{url}
\usepackage{graphicx}
\usepackage{natbib}
\usepackage{caption}
\usepackage{algorithm}
\usepackage{algorithmic}
\usepackage{booktabs}
\usepackage{array}
\usepackage{multirow}
\usepackage{tabularx}
\usepackage{amsmath}
\usepackage{amssymb}

\title{Learning from Think-Mode Advantage via On-Policy Distillation}
\author{
    Wanqi Ren\thanks{Work done during an internship at ByteDance.}\textsuperscript{\rm 1},
    Jianxiang Wang\textsuperscript{\rm 1},
    Danxuan Liu\textsuperscript{\rm 1},\\
    Linyi Ding\textsuperscript{\rm 1},
    Huaixiao Tou\textsuperscript{\rm 1}
}
\affiliations{
    \textsuperscript{\rm 1}ByteDance, China\\
    renwanqi@bytedance.com
}

\newcommand{\method}{ThinkOPD}
\newcommand{\vanillamethod}{Uniform ThinkOPD}
\newcommand{\uniformabbr}{U-ThinkOPD}
\newcommand{\opsdmethod}{OPSD}
\newcommand{\student}{\pi_S}
\newcommand{\teacher}{\pi_T}
\newcommand{\trd}{\mathrm{TRD}}
\newcommand{\avgk}{avg@16}
\newcommand{\bestk}{best@16}
\newcommand{\secondeval}[1]{\underline{#1}}
\newcommand{\contextgroup}[1]{\begin{tabular}{@{}c@{}}#1\end{tabular}}

\newcommand{\methodcell}[1]{#1}
\newcommand{\metrichead}[1]{#1}

\begin{document}

\maketitle

\begin{abstract}
Explicit intermediate reasoning gives large language models (LLMs) a stronger problem-solving mode.  We study learning from this think-mode advantage via on-policy distillation (OPD).  OPD preserves student-generated trajectories and provides dense token-level teacher targets at student-visited prefixes.  Privileged reasoning is used during distillation rather than student inference.  \vanillamethod{}, a natural think-enabled OPD baseline, conditions a fixed teacher on one shared think trace and uniformly distills every sibling student response.  Although its prefixes are on-policy, the trace need not follow a route compatible with every complete response: the same privileged trace can induce different teacher--student discrepancies even when responses reach the same outcome.  We summarize this interaction with \emph{trace--response divergence} (TRD) and introduce \method{}, which routes supervision at the response level by combining group-relative reward gain with a TRD-based compatibility proxy.  Final response weights are normalized within each rollout group.  Across mathematical reasoning and code generation, \method{} outperforms \vanillamethod{} in both same-model settings and both cross-model teacher--student pairs, and it exceeds representative rationale and self-distillation baselines in a controlled comparison.  Controlled interventions show that outcome benefit and the TRD-based proxy provide complementary routing signals in this setting.  Think-enabled OPD provides a controlled setting for studying how teacher advantage becomes transferable along student responses.
\end{abstract}

\section{Introduction}

Intermediate reasoning helps LLMs solve problems that resist direct generation~\citep{wei2022cot,wang2022selfconsistency}.  Recent interfaces expose this capability through explicit think and no-think modes~\citep{yang2025qwen3}.  We study learning from this think-mode advantage via OPD: a no-think student explores its own trajectories while a think-mode teacher supplies privileged supervision during distillation.  We define no-think inference as generation without conditioning on the teacher's privileged think trace, without constraining response length.  The central question is how effectively the stronger think-mode view transfers through the sequence of states that the student visits.

Several transfer routes are possible.  Offline rationale tuning and reasoning internalization learn from completed teacher solutions but anchor supervision to fixed target sequences~\citep{hsieh2023distilling,ho2023reasoning,li2023scotd,yu2024system2,xu2025twt}.  Reinforcement learning preserves student exploration but typically communicates task success through outcome-level rewards~\citep{yu2025dapo}.  Self-distillation offers a third route, and OPD is particularly suitable here: it samples trajectories from the current student, then queries dense teacher distributions at the states that the student actually visits~\citep{gu2024minillm,agarwal2024opd,ye2026opcd}.  Our matched baselines instantiate this space differently: SDFT uses a demonstration-conditioned self-teacher to produce on-policy signals, OPSD uses a fixed privileged self-teacher, SDPO updates supervision with feedback, and BRTS searches over multiple teacher trajectories~\citep{shenfeld2026sdft,zhao2026opsd,hubotter2026sdpo,zhang2026brts}.  We study privileged self-distillation under think-enabled OPD, pairing student exploration with a stronger teacher view.

A direct construction conditions a fixed teacher on its privileged think trace and uniformly distills the current student's rollouts.  We call this controlled baseline \vanillamethod{}.  Throughout, \emph{think-enabled OPD} denotes the broader setting.  At first glance, the baseline offers both a stronger teacher view and on-policy state coverage.  These properties concern different objects: the scored prefix comes from the student, but the privileged trace follows the teacher's solution route.  The same trace can suit one response and conflict with another even when their outcomes are identical.

\begin{figure*}[t]
\centering
\includegraphics[width=\textwidth]{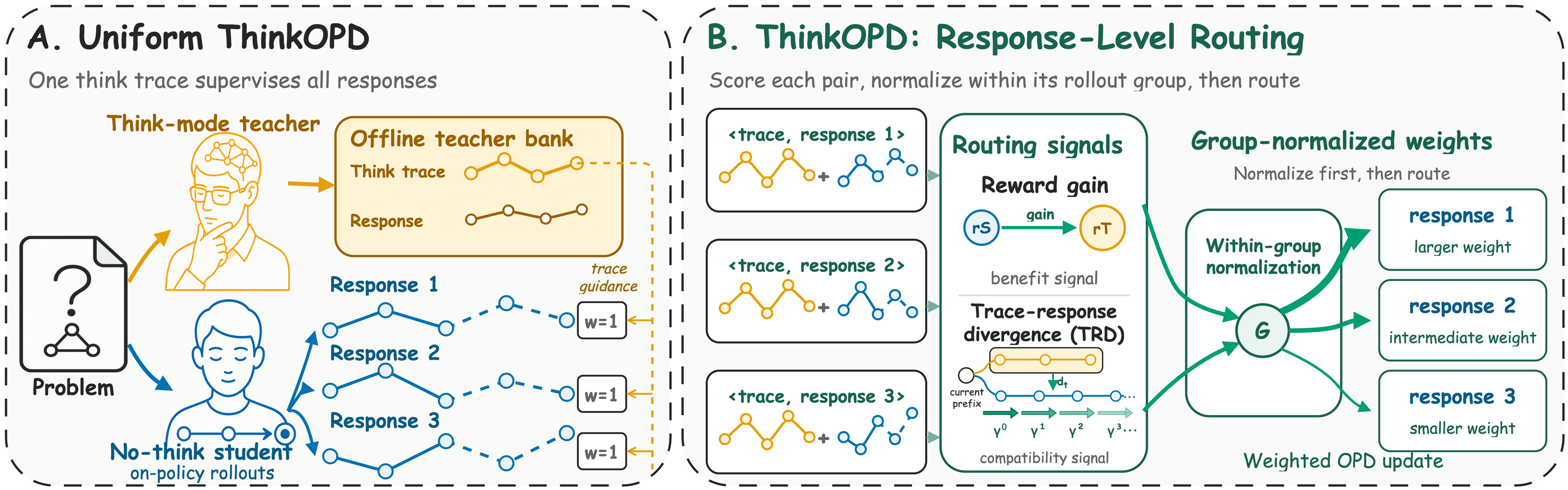}
\caption{\method{} response routing under a shared privileged trace.  \vanillamethod{} weights sibling responses equally; \method{} routes them using outcome benefit and a TRD-based compatibility proxy with group-normalized weights.}
\label{fig:paradigm_schematic}
\end{figure*}

The natural think-enabled OPD construction reveals a trace--response discrepancy while holding the teacher model fixed.  This makes learning from think-mode advantage a controlled setting for studying how outcome benefit and trace--response compatibility interact during transfer.  Prior comparisons often vary teacher capacity, family, checkpoint, or data together with compatibility~\citep{mirzadeh2020teacherassistant,xu2025strongerteachers,li2025smallmodels}.  Here the think trace instead exposes a concrete solution route whose compatibility can vary across complete student responses.  We summarize this interaction with \emph{trace--response divergence} (TRD), an operational response-level proxy for compatibility.

The conflict suggests two complementary decisions.  Reward gain measures \emph{outcome benefit}: whether the teacher obtains a better outcome for a response.  TRD operationalizes \emph{trace--response compatibility} through the discrepancy induced by the shared trace along that response.  Neither signal alone defines distillation utility.  Low discrepancy without benefit can emphasize an already adequate response, while benefit without a compatibility proxy can favor supervision delivered through an unsuitable route.  \method{} treats each trace--response pair as a routing unit (Figure~\ref{fig:paradigm_schematic}).  Within each prompt group, reward gain ranks outcome benefit, and inverse relative TRD from discounted future KL ranks responses by this compatibility proxy.  Their product is normalized within the group to obtain the response-level OPD weight.  Because \vanillamethod{} and \method{} share the teacher view and OPD interface, their comparison isolates sibling-response routing (Table~\ref{tab:method_interfaces}).

Across both same-model settings, \method{} improves five-benchmark Overall \avgk{} over \vanillamethod{} by \(1.7\)--\(3.0\) percentage points.  It also improves pooled math \avgk{} by \(1.6\)--\(1.9\) points in both cross-model pairs.  On Qwen3-1.7B, it exceeds the highest-scoring external baseline by \(2.3\) points.  The controlled component analysis directly tests the preceding design claim: TRD-only routing falls below uniform weighting, reward-only routing remains below the full method, their combination performs best, and reversing the TRD preference is harmful.  Thus, reward gain captures outcome benefit, while lower TRD favors routes that are more compatible under the proposed proxy.

Our contributions are threefold:
\begin{itemize}
    \item We identify TRD in natural think-enabled OPD, revealing a controlled setting for studying how outcome benefit and trace--response compatibility interact during transfer.
    \item We define TRD over complete student solution paths and propose \method{}, a group-relative router combining reward gain with a TRD-based compatibility proxy.
    \item We show consistent same- and cross-model gains and controlled evidence that outcome benefit and trace compatibility are complementary.
\end{itemize}

\section{Related Work}

\begin{table}[!t]
\centering
\small
\setlength{\tabcolsep}{1.5pt}
\renewcommand{\arraystretch}{1.08}
\newcommand{\yesmark}{\(\checkmark\)}
\newcommand{\nomark}{\(\times\)}
\begin{tabular*}{\columnwidth}{@{\extracolsep{\fill}}lcccc@{}}
\toprule
Method &
On-policy &
Compatibility &
Benefit &
Multi-Trace \\
\midrule
SDFT & \yesmark & \nomark & \nomark & \nomark \\
OPSD/SDPO & \yesmark & \nomark & \nomark & \nomark \\
BRTS & \yesmark & \yesmark & \yesmark & \yesmark \\
\uniformabbr{} & \yesmark & \nomark & \nomark & \nomark \\
\textbf{\method{}} & \yesmark & \yesmark & \yesmark & \nomark \\
\bottomrule
\end{tabular*}
\caption{Supervision interfaces compared by on-policy scoring, compatibility modeling, outcome-aware selection or weighting, and use of multiple teacher traces.  Multi-Trace means that a method generates or compares multiple teacher reasoning trajectories per training instance; it does not denote sibling student responses.  All methods use privileged teacher context, and U-ThinkOPD denotes Uniform ThinkOPD.}
\label{tab:method_interfaces}
\end{table}

\paragraph{Reasoning transfer and privileged self-distillation.}
Sequence and rationale distillation use fixed teacher targets, while reinforcement learning uses task feedback~\citep{hinton2015distilling,kim2016sequence,hsieh2023distilling,ho2023reasoning,yu2024system2,xu2025twt,yu2025dapo}.  OPD and MiniLLM align teachers to student trajectories.  Speculative KD interleaves student proposals and teacher corrections under unreliable feedback~\citep{agarwal2024opd,gu2024minillm,xu2025speculativekd}.  OPCD adds context-conditioned teachers~\citep{ye2026opcd}.  Our controls span demonstration-conditioned on-policy SDFT, fixed privileged OPSD, and feedback-conditioned SDPO~\citep{shenfeld2026sdft,zhao2026opsd,hubotter2026sdpo}.  ThinkOPD fixes the teacher view and routes one trace across sibling responses, separating response routing from teacher evolution and rationale generation.

\paragraph{Compatibility and selective supervision.}
Teacher strength can fail under capacity or reasoning mismatch~\citep{mirzadeh2020teacherassistant,xu2025strongerteachers,li2025smallmodels,li2026rethinking}.  Selective objectives adapt local supervision by entropy, teachability, position, or near-future guidance~\citep{jin2026entropyaware,wang2026teachability,liu2026pwopsd,jiang2026topd}.  Concurrent BRTS instead searches teacher trajectories~\citep{zhang2026brts}.  ThinkOPD aggregates discrepancy across each response and routes a shared trace by outcome benefit and compatibility.

\FloatBarrier

\section{Method}

\method{} treats each shared-trace/student-response pair as the unit of supervision.  It first estimates response-level trace compatibility and then combines it with outcome benefit to route dense OPD targets across sibling trajectories.

\subsection{Problem Formulation}

We consider a no-think student \(\student\) and a fixed think-mode teacher \(\teacher\).  During offline bank construction, the teacher jointly generates each think trace \(z\) and a paired response \(\widetilde y_T\).  A task verifier scores the response before the trace--response--reward record is stored.  At training time, the current student samples \(G\) sibling responses \(y^{(1)},\ldots,y^{(G)}\) for prompt \(x\).  The resulting prompt group \(\mathcal{G}(x)\) holds \(x\) and the selected \(z\) fixed while varying the student trajectory.  The teacher-side record paired with sibling \(k\) supplies \(r_T^{(k)}=r(\widetilde y_T^{(k)})\), and the online student response receives \(r_S^{(k)}=r(y^{(k)})\).  At prefix \(y_{<t}^{(k)}\), the two policy views are:
\begin{equation}
\begin{aligned}
    p_{S,t}^{(k)}(\cdot)
    &= \student(\cdot \mid x,y_{<t}^{(k)}), \\
    p_{T,t}^{z,(k)}(\cdot)
    &= \teacher(\cdot \mid x,z,y_{<t}^{(k)}).
\end{aligned}
\label{eq:teacher_mode_views}
\end{equation}
The student samples without \(z\).  The teacher conditions on \(z\) only when scoring the same prefix.  We compute sparse forward KL on the teacher's top-\(K\) support, with \(q_{T,t}^{(k)}\) and \(q_{S,t}^{(k)}\) denoting the corresponding renormalized views.  Because \(z\) is fixed within a prompt group, we suppress it in this notation below.  We denote by \(\mathcal{R}(y^{(k)})\) all valid generated-token positions in the complete student response, excluding prompt and padding tokens.  Every sampled response and every valid token within it participate in scoring.  Here \(G\) determines the number of sibling trajectories in each within-prompt comparison set, and \(K\) determines the sparse teacher support used to evaluate token-level disagreement.  Both are fixed by the training protocol.

\subsection{\vanillamethod{}}

To isolate response routing, \vanillamethod{} and \method{} use the same offline trace--response--reward bank, student rollouts, and response-conditioned teacher scores.  The teacher view \(q_{T,t}^{(k)}\) is queried at every student prefix and is response-specific.  The paired teacher response stored in the bank supplies \(r_T^{(k)}\).

Let \(\mathcal{V}\) be the scored response-token triples in an update.  The direct baseline, \vanillamethod{}, gives every trace--response pair the same OPD weight:
\begin{equation}
    \mathcal{L}_{\mathrm{vanilla}}
    = \frac{1}{|\mathcal{V}|}
    \sum_{(x,k,t)\in\mathcal{V}}
    \mathrm{KL}\!\left(q_{T,t}^{(k)}\,\Vert\,q_{S,t}^{(k)}\right).
\label{eq:vanilla_thinkopd}
\end{equation}
This is prefix-level on-policy supervision: every target is queried at a state visited by the current student.

\subsection{Trace--Response Divergence}

The on-policy property of Equation~\ref{eq:vanilla_thinkopd} applies to the prefix, not to the privileged trace.  Because \(z\) follows the teacher's solution route, reusing it across siblings couples one teacher trajectory with several student trajectories.  \vanillamethod{} nevertheless treats every coupling as equally useful.  TRD measures the discrepancy induced by trace-conditioned supervision as each complete student response unfolds.

Holding teacher capacity, family, and checkpoint fixed makes route-dependent discrepancy directly observable at the trace--response level, providing a controlled compatibility proxy for think-enabled OPD.

\begin{figure}[t]
\centering
\includegraphics[width=\columnwidth]{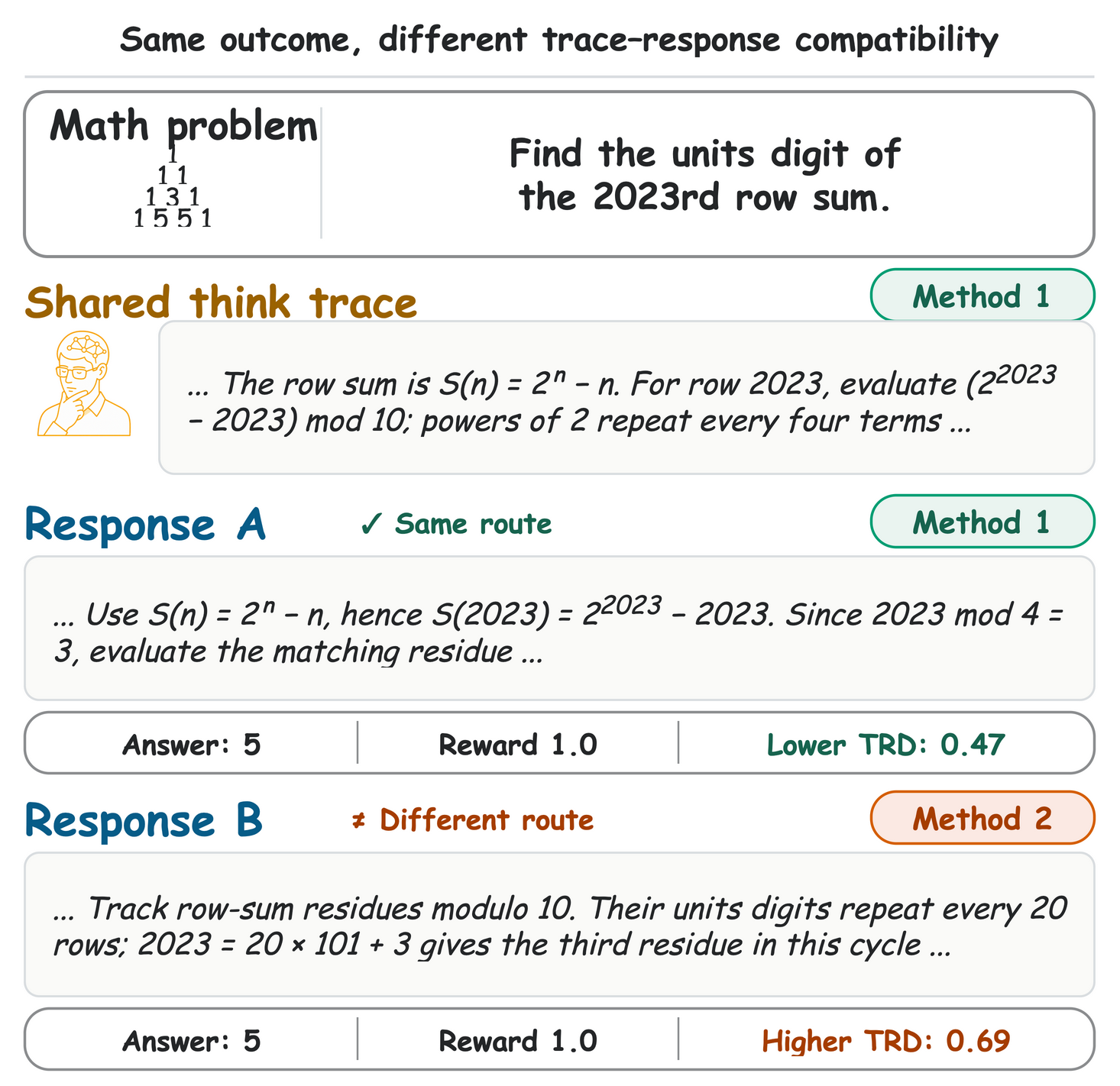}
\caption{A controlled trace--response discrepancy case.  One closed-form think trace conditions two correct responses with the same answer and reward; the route-aligned response has lower TRD than the alternative recurrence route.}
\label{fig:trd_schematic}
\end{figure}

Figure~\ref{fig:trd_schematic} illustrates the interaction while holding the problem, teacher trace, answer, and reward fixed.  The response following the teacher's closed-form route has lower TRD than the alternative correct route.  Equal outcomes can conceal different trace-conditioned discrepancies.

TRD captures a sequential interaction rather than a single-token discrepancy.  A local KL spike can reflect a temporary lexical choice, whereas a route-level departure can alter teacher targets across several subsequent prefixes.  Discounted continuation distinguishes these patterns by accumulating persistent disagreement while emphasizing its nearer consequences.  The finite horizon bounds this continuation, and response aggregation yields one operational score for the complete trace--response pair.

We construct TRD in three stages: local disagreement, discounted continuation, and response aggregation.  For response \(k\), sparse forward KL first records the trace-conditioned disagreement at position \(t\); a bounded future sum then captures whether that discrepancy persists; averaging these local-future values produces one response-level score.  Formally:
\begin{align}
    D_t^{(k)} &=
    \mathrm{KL}\!\left(q^{(k)}_{T,t}\,\Vert\,q^{(k)}_{S,t}\right)
\label{eq:token_forward_kl}
\\
    \trd_t^{(k)}
    &= \sum_{\tau=t}^{\min\{t+H,\lvert y^{(k)}\rvert\}}
    \gamma^{\tau-t}D_{\tau}^{(k)}m_{\tau}^{(k)},
    \qquad 0<\gamma<1
\label{eq:local_trd}
\\
    \overline{\trd}^{(k)} &=
    \frac{1}{|\mathcal{R}(y^{(k)})|}
    \sum_{t\in\mathcal{R}(y^{(k)})}\trd_t^{(k)}
\label{eq:response_trd}
\end{align}
Here \(m_{\tau}^{(k)}\in\{0,1\}\) masks invalid generated positions, \(H\) bounds the continuation window, and \(0<\gamma<1\) discounts more distant discrepancies.  The set \(\mathcal{R}(y^{(k)})\) contains the valid positions used for response aggregation.  Low \(\overline{\trd}^{(k)}\) means trace-conditioned targets remain close to the student across the response.  High values indicate sustained departure.  This aggregation turns token-level KL into a response-level trace--response compatibility proxy.

\subsection{Group-Relative Response Routing}

\begin{figure}[t]
\centering
\includegraphics[width=\columnwidth]{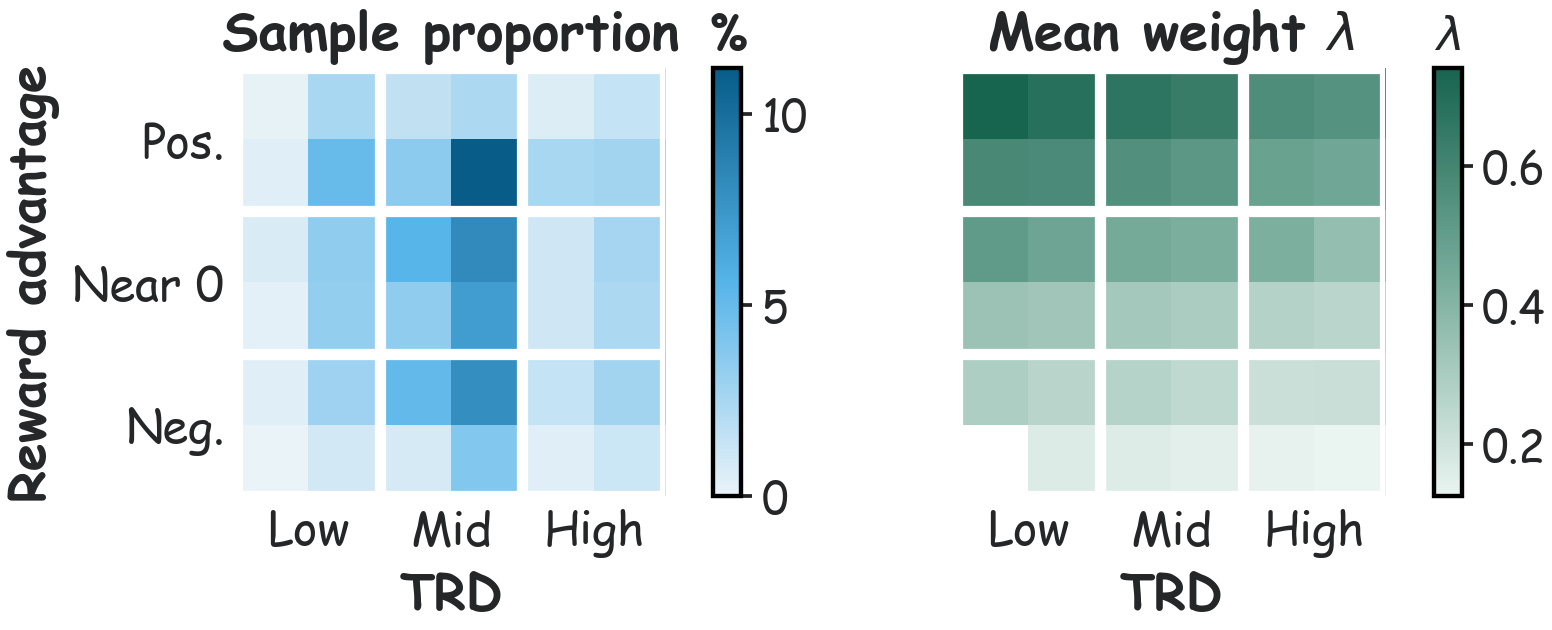}
\caption{Sample mass (left) and mean response weight (right) for 768 valid responses in 192 four-response prompt groups from Qwen3-1.7B, binned by reward advantage and TRD.  Each coarse region spans two fine bins per axis.}
\label{fig:utility_regimes}
\end{figure}

Having measured the shared-trace interaction, \method{} routes supervision with two signals.  Reward gain compares response-specific teacher and student rewards, and TRD measures how the shared trace interacts with each sibling trajectory.  Because both vary across prompts, we compare them only within the same prompt group.  Let \(\operatorname{Std}_{\mathcal{G}}\) denote z-score standardization across siblings in \(\mathcal{G}(x)\):
\begin{equation}
\begin{aligned}
    g^{(k)} &= r_T^{(k)}-r_S^{(k)}, \\
    \widehat g^{(k)}
    &=\operatorname{Std}_{\mathcal{G}}\!\left(g^{(k)}\right), \\
    \widehat{\trd}^{(k)}
    &=\operatorname{Std}_{\mathcal{G}}\!\left(\overline{\trd}^{(k)}\right).
\end{aligned}
\label{eq:relative_routing_signals}
\end{equation}
Because both rewards are response-specific, \(\widehat g^{(k)}\) retains teacher-side as well as student-side variation within the group.  Larger \(\widehat g^{(k)}\) indicates greater outcome benefit, and smaller \(\widehat{\trd}^{(k)}\) indicates greater compatibility under the operational proxy.  Their response score is:
\begin{equation}
\begin{aligned}
    \widetilde{\lambda}^{(k)}
    &= \sigma\!\left(\widehat g^{(k)}\right)
       \exp\!\left[-\eta\,\sigma\!\left(\widehat{\trd}^{(k)}\right)\right],\\
    \lambda^{(k)}
    &=\widetilde{\lambda}^{(k)}/Z_x .
\end{aligned}
\label{eq:response_utility}
\end{equation}
Here \(\sigma\) is the sigmoid function.  The coefficient \(\eta>0\) controls routing contrast: larger values suppress relatively high-TRD responses more strongly, while smaller values approach benefit-dominated routing.  The group-specific normalizer \(Z_x=G^{-1}\sum_{j=1}^{G}\widetilde{\lambda}^{(j)}\) preserves the response ranking.  The first factor increases with relative outcome benefit, and the second decreases smoothly with relative TRD without introducing a hard threshold.  Figure~\ref{fig:utility_regimes} visualizes this joint routing pattern across reward-advantage and TRD bins and reports the sample mass in every region of the grid.

The two normalization stages make routing relational within each prompt group.  Each response is compared only with siblings generated for the same problem and paired with the same trace; the resulting response weights are then normalized within that group.  Consequently, \(\lambda^{(k)}\) reallocates dense OPD supervision among sibling paths according to their relative benefit and TRD.

The routing parameters govern complementary decisions: \(H\) sets how far the trace--response interaction is observed, while \(\eta\) controls its effect on response weights.  We examine both with the within-window discount \(\gamma\) fixed.

Figure~\ref{fig:utility_regimes} connects the motivating phenomenon to the router.  Sibling responses occupy low, middle, and high TRD bins even within the same reward-advantage band, so outcome benefit does not determine the trace-conditioned interaction.  The weight map assigns its largest coefficients where positive benefit coincides with low TRD; at comparable benefit, weight falls as TRD increases.  The two signals therefore organize distinct axes of the response-level decision.

\subsection{Training Objective and Procedure}

Each response coefficient is shared by all valid generated-token positions and fixed during the policy update.  With \(\operatorname{sg}(\cdot)\) denoting stop-gradient, the weighted objective is:
\begin{equation}
    \mathcal{L}_{\mathrm{ThinkOPD}} =
    \frac{1}{|\mathcal{V}|}
    \sum_{(x,k,t)\in\mathcal{V}}
    \operatorname{sg}\!\left(\lambda^{(k)}\right)
    \mathrm{KL}\!\left(q^{(k)}_{T,t}\,\Vert\,q^{(k)}_{S,t}\right),
\label{eq:thinkopd_objective}
\end{equation}
where \(\mathcal{V}\) is the same set of scored positions as in Equation~\ref{eq:vanilla_thinkopd}.  Setting every \(\lambda^{(k)}\) to one recovers \vanillamethod{}.

Sharing \(\lambda^{(k)}\) across the response keeps the routing decision aligned with the complete trace--response pair from which TRD is computed.  The underlying supervision remains token-level and dense, but its aggregate contribution reflects whether that student path offers both outcome benefit and compatible trace-conditioned guidance.  This separates the granularity of supervision from the granularity of routing.

At each step, the student generates sibling responses before routing, and the teacher scores their visited prefixes under the shared trace.  The offline bank supplies fixed context and teacher records, while current paths receive recomputed response-conditioned targets.  This preserves the defining OPD interaction summarized in Algorithm~\ref{alg:thinkopd}.

\begin{algorithm}[t]
\caption{\method{}}
\label{alg:thinkopd}
\small
\textbf{Input:} Offline trace--response--reward bank, student \(\student\), teacher \(\teacher\),
and routing parameters\\
\textbf{Output:} Updated no-think student policy \(\student\)
\begin{algorithmic}[1]
\FOR{each minibatch \(\mathcal{X}\) of prompts}
    \FOR{each prompt \(x\in\mathcal{X}\)}
        \STATE Retrieve the shared trace \(z\) and paired teacher responses/rewards
        \STATE Sample sibling responses \(y^{(1:G)}\sim\student(\cdot\mid x)\)
        \FOR{each response \(y^{(k)}\)}
            \STATE Score \(y^{(k)}\) and compute response-token KL \(D_t^{(k)}\)
            \STATE Compute reward gain \(g^{(k)}\) and response TRD \(\overline{\trd}^{(k)}\)
        \ENDFOR
        \STATE Standardize reward gain and TRD within \(\mathcal{G}(x)\)
        \STATE Compute and stop-gradient \(\lambda^{(k)}\) with Equation~\ref{eq:response_utility}
    \ENDFOR
    \STATE Update \(\student\) with weighted full-response forward KL
\ENDFOR
\end{algorithmic}
\end{algorithm}
\FloatBarrier

\begin{samepage}
\section{Experiments}

We evaluate \method{} across model scales, mathematical reasoning, code generation, and cross-model distillation.  The experiments answer four questions:

\noindent\textbf{Q1} Does routing by outcome benefit and a TRD-based compatibility proxy improve think-mode transfer?

\noindent\textbf{Q2} Does it generalize across models, tasks, and model pairs?

\noindent\textbf{Q3} Does the TRD proxy outperform alternative compatibility proxies?

\noindent\textbf{Q4} Is response routing more efficient than trajectory search?
\end{samepage}

\subsection{Settings}

\noindent\textbf{Models.}
The primary comparison uses Qwen3-1.7B as both student and think-mode teacher.  Same-model experiments additionally evaluate Qwen3-0.6B~\citep{yang2025qwen3}; cross-model experiments pair Qwen3-4B-Thinking-2507 with Qwen3-1.7B and Qwen3-1.7B with Qwen3-0.6B.

\noindent\textbf{Training data and evaluation benchmarks.}
Training and evaluation data are disjoint in both domains.  Math training uses DAPO-Math-17K~\citep{yu2025dapo}, with 256 prompts held out for checkpoint selection; evaluation uses AIME24, AIME25, and AMC23~\citep{huggingfaceh4aime2024,opencompassaime2025,washbourne2025amc23}.  Code training uses Skywork-OR1-Coding-14K~\citep{he2025skywork} with a separate 256-problem validation split; evaluation uses Skywork heldout512 and LiveCodeBench v5~\citep{jain2025livecodebench}.  We report \avgk{}, the mean correctness over 16 samples per problem, and \bestk{}, the fraction of problems solved at least once.

\noindent\textbf{Baselines.}
\vanillamethod{} is the uniform-routing control: it shares the fixed think-mode teacher, offline bank, student rollouts, training budget, and checkpoint-selection protocol with \method{}, but assigns equal weight to every response.  Cross-model experiments use the same control.

The method families motivating the external comparison are summarized in Table~\ref{tab:method_interfaces}.  On the representative Qwen3-1.7B setting, SDFT, OPSD, SDPO, and BRTS use matched training data, response budgets, checkpoint selection, and evaluation.  Detailed implementations and protocol adaptations appear in the supplementary material.

\noindent\textbf{Implementation.}
\method{} and its uniform control use full-parameter BF16 training with verl~\citep{sheng2025hybridflow} for 100 steps, batch size 64, and learning rate \(10^{-6}\) on 32 high-performance GPUs.  Teacher distributions are recomputed on sampled student prefixes, and checkpoints are selected by held-out validation avg@3.  The primary Qwen3-1.7B math comparison averages three independent runs.  Other model-task settings, baselines, interventions, proxies, and sensitivity points are matched validation-selected point estimates.  The supplement provides full implementation, uncertainty, and resource details.

\noindent\textbf{Offline Teacher Bank and Asynchronous OPD.}
To reduce training latency, we use two complementary measures.  First, an immutable offline Teacher Bank stores the fixed teacher's think trace, paired response, verifier reward, and provenance for each prompt.  The student still samples fresh responses, and the teacher recomputes forward-only distributions at visited prefixes, so scoring remains on-policy.  Second, asynchronous OPD overlaps rollout, trace-conditioned teacher scoring, and optimization through a bounded queue of complete, single-version sibling groups; consumed groups lag by at most one update.  The bank removes repeated teacher generation, while asynchronous execution reduces pipeline idle time.  Further details appear in the supplementary material.

\subsection{Results and Analysis}

\subsubsection{ThinkOPD Outperforms Matched Baselines}

\begin{table}[t]
\centering
\begingroup
\small
\setlength{\tabcolsep}{0.6pt}
\renewcommand{\arraystretch}{1.04}
\begin{tabular*}{\columnwidth}{@{\extracolsep{\fill}}l*{6}{c}@{}}
\toprule
\multirow{2}{*}{Method} & \multicolumn{3}{c}{Math} & \multicolumn{2}{c}{Coding} & \multirow{2}{*}{Overall} \\
\cmidrule(lr){2-4}\cmidrule(lr){5-6}
& \metrichead{AIME24} & \metrichead{AIME25} & \metrichead{AMC23} & \metrichead{Skywork} & \metrichead{LCBv5} & \\
\midrule
SDFT & 28.4 & \secondeval{25.2} & 56.9 & 18.8 & 12.0 & 28.2 \\
\opsdmethod{} & 28.8 & 24.7 & 57.7 & 19.4 & 13.0 & 28.7 \\
SDPO & 30.0 & 25.1 & \secondeval{58.4} & 20.2 & \textbf{13.5} & 29.4 \\
BRTS & \secondeval{30.1} & \textbf{26.6} & 58.2 & \secondeval{20.4} & \secondeval{13.4} & \secondeval{29.7} \\
\mbox{\uniformabbr{}} & 29.2 & 25.0 & 57.6 & 20.1 & 13.3 & 29.0 \\
\methodcell{\method{}} & \methodcell{\textbf{35.2}} & \methodcell{\secondeval{25.4}} & \methodcell{\textbf{60.8}} & \methodcell{\textbf{25.2}} & \methodcell{\secondeval{13.4}} & \methodcell{\textbf{32.0}} \\
\bottomrule
\end{tabular*}
\endgroup
\caption{Qwen3-1.7B baseline comparison under eval16 (\%, \avgk{}; higher is better).  Overall is the unweighted mean over the five benchmarks.  Bold and underlined entries indicate first and second place.}
\label{tab:baseline_comparison}
\end{table}

\begin{table*}[t]
\centering
\begingroup
\small
\setlength{\tabcolsep}{1.5pt}
\renewcommand{\arraystretch}{1.10}
\begin{tabularx}{\textwidth}{@{}>{\centering\arraybackslash}p{0.105\textwidth}>{\raggedright\arraybackslash}p{0.145\textwidth}*{10}{>{\centering\arraybackslash}X}@{\hspace{2.5pt}}*{2}{>{\centering\arraybackslash}X}@{}}
\toprule
\multirow{3}{*}{Model} & \multirow{3}{*}{Method} & \multicolumn{6}{c}{Math} & \multicolumn{4}{c}{Coding} & \multicolumn{2}{c}{\multirow{2}{*}{Overall}} \\
\cmidrule(lr){3-8}\cmidrule(lr){9-12}
 & & \multicolumn{2}{c}{AIME24} & \multicolumn{2}{c}{AIME25} & \multicolumn{2}{c}{AMC23} & \multicolumn{2}{c}{Skywork held.} & \multicolumn{2}{c}{LCBv5} & \multicolumn{2}{c}{} \\
\cmidrule(lr){3-4}\cmidrule(lr){5-6}\cmidrule(lr){7-8}\cmidrule(lr){9-10}\cmidrule(lr){11-12}\cmidrule(lr){13-14}
 & & \metrichead{Avg.} & \metrichead{Best} & \metrichead{Avg.} & \metrichead{Best} & \metrichead{Avg.} & \metrichead{Best} & \metrichead{Avg.} & \metrichead{Best} & \metrichead{Avg.} & \metrichead{Best} & \metrichead{Avg.} & \metrichead{Best} \\
\midrule
\multirow{2}{*}{\contextgroup{Qwen3 0.6B}} & Think off & 2.5 & 16.6 & 2.5 & 6.6 & 23.5 & 62.6 & 10.1 & 24.8 & 6.1 & 15.0 & 8.9 & 25.1 \\
 & Think on & 8.7 & 36.6 & 17.0 & 40.0 & 46.3 & 78.3 & 19.5 & 41.7 & 12.8 & 24.7 & 20.9 & 44.3 \\
\cmidrule(lr){2-14}
\multirow{2}{*}{\contextgroup{Qwen3 1.7B}} & Think off & 13.5 & 46.6 & 10.0 & 30.0 & 40.6 & 73.4 & 18.1 & 33.3 & 11.2 & 21.8 & 18.6 & 41.0 \\
 & Think on & 43.3 & 76.6 & 36.0 & 70.0 & 73.7 & 92.7 & 40.1 & 58.0 & 31.1 & 46.2 & 44.8 & 68.7 \\
\midrule
\multirow{2}{*}{\contextgroup{Qwen3 0.6B}} & \uniformabbr{} & \textbf{2.2} & \textbf{16.6} & 3.3 & 10.0 & 23.7 & 61.4 & 16.8 & 39.4 & 10.5 & 22.5 & 11.3 & 29.9 \\
& \methodcell{\method{}} & \methodcell{1.6} & \methodcell{13.3} & \methodcell{\textbf{4.5}} & \methodcell{\textbf{20.0}} & \methodcell{\textbf{25.9}} & \methodcell{\textbf{68.6}} & \methodcell{\textbf{19.7}} & \methodcell{\textbf{42.5}} & \methodcell{\textbf{13.7}} & \methodcell{\textbf{25.8}} & \methodcell{\textbf{13.0}} & \methodcell{\textbf{34.0}} \\
\cmidrule(lr){2-14}
\multirow{2}{*}{\contextgroup{Qwen3 1.7B}} & \uniformabbr{} & 29.2 & 65.5 & 25.0 & \textbf{62.2} & 57.6 & \textbf{88.3} & 20.1 & 41.4 & 13.3 & 27.9 & 29.0 & 57.0 \\
& \methodcell{\method{}} & \methodcell{\textbf{35.2}} & \methodcell{\textbf{73.3}} & \methodcell{\textbf{25.4}} & \methodcell{61.1} & \methodcell{\textbf{60.8}} & \methodcell{87.5} & \methodcell{\textbf{25.2}} & \methodcell{\textbf{55.0}} & \methodcell{\textbf{13.4}} & \methodcell{\textbf{41.2}} & \methodcell{\textbf{32.0}} & \methodcell{\textbf{63.6}} \\
\bottomrule
\end{tabularx}
\endgroup
\caption{Same-model performance under eval16 across mathematical reasoning and code generation (\%, higher is better).  Overall Avg.\ and Best are the unweighted means of their corresponding five benchmark columns.  Think-off and think-on rows are undistilled base-policy references; within each distilled model group, boldface marks the higher result.}
\label{tab:main}
\end{table*}

\begin{table}[t]
\centering
\begingroup
\small
\newcommand{\crosspairgroup}[2]{\begin{tabular}{@{}c@{}}#1\\[-1pt]$\downarrow$\\[-1pt]#2\end{tabular}}
\newcommand{\crossmethod}[1]{#1}
\setlength{\tabcolsep}{0.2pt}
\renewcommand{\arraystretch}{1.08}
\begin{tabular*}{\columnwidth}{@{\extracolsep{\fill}}>{\centering\arraybackslash}p{0.10\columnwidth}>{\raggedright\arraybackslash}p{0.20\columnwidth}*{5}{>{\centering\arraybackslash}p{0.134\columnwidth}}@{}}
\toprule
Model & Method & \metrichead{AIME24} & \metrichead{AIME25} & \metrichead{AMC23} & \metrichead{Avg.} & \(\Delta\) \\
\midrule
\multirow{3}{*}{\crosspairgroup{4B}{1.7B}} & \crossmethod{OPD} & 33.9 & 29.7 & 66.2 & 51.8 & 0.0 \\
 & \crossmethod{\mbox{\uniformabbr{}}} & 35.2 & 30.6 & 65.5 & 51.8 & 0.0 \\
& \methodcell{\crossmethod{\method{}}} & \methodcell{\textbf{37.5}} & \methodcell{\textbf{34.3}} & \methodcell{\textbf{66.5}} & \methodcell{\textbf{53.7}} & \methodcell{\textbf{+1.9}} \\
\midrule
\multirow{3}{*}{\crosspairgroup{1.7B}{0.6B}} & \crossmethod{OPD} & 2.5 & 1.4 & 24.5 & 15.0 & -6.0 \\
 & \crossmethod{\mbox{\uniformabbr{}}} & \textbf{5.2} & 6.6 & 31.9 & 21.0 & 0.0 \\
& \methodcell{\crossmethod{\method{}}} & \methodcell{5.0} & \methodcell{\textbf{7.0}} & \methodcell{\textbf{34.5}} & \methodcell{\textbf{22.6}} & \methodcell{\textbf{+1.6}} \\
\bottomrule
\end{tabular*}
\endgroup
\caption{Cross-model Qwen3 math performance under eval16 (\%, \avgk{}, higher is better); the first column gives teacher and student sizes.  Avg.\ pools all AIME24, AIME25, and AMC23 problems, and \(\Delta\) is its gain over \vanillamethod{} within each pair.}
\label{tab:cross_model}
\end{table}

Table~\ref{tab:baseline_comparison} tests the central claim under matched supervision.  Relative to \vanillamethod{}, \method{} raises five-benchmark Overall \avgk{} from \(29.0\) to \(32.0\).  This \(3.0\)-point difference isolates response-level routing in our implementation because the teacher, trace bank, student rollouts, and optimization protocol are unchanged.  \method{} also exceeds the strongest external baseline, BRTS at \(29.7\), by \(2.3\) points.

The external baselines occupy a narrow Overall range of \(28.2\)--\(29.7\), and \vanillamethod{} is already competitive within it.  This comparison sharpens the motivating distinction.  A privileged think trace supplies potential outcome benefit, but its availability does not determine which sibling response can use that supervision coherently.  The matched improvement from \vanillamethod{} to \method{} is consistent with routing this shared signal by both benefit and trace--response compatibility.

\method{} achieves the strongest Overall result through clear gains on AIME24, AMC23, and Skywork while remaining competitive on AIME25 and LCBv5.  This breadth distinguishes response routing from baselines whose improvements concentrate on individual benchmarks.

\subsubsection{Gains Across Models and Tasks}

Table~\ref{tab:main} separates teacher-side opportunity from realized transfer.  Enabling think mode creates \(12.0\)--\(26.2\) points of base-policy headroom across the two Qwen3 scales, but this headroom is an opportunity rather than the amount that uniform distillation can recover.  Across both same-model scales, \method{} improves Overall by \(1.7\)--\(3.0\) points over the uniform control.

The two cross-model settings make the distinction more visible.  For Qwen3-4B\(\rightarrow\)1.7B, no-think OPD and uniform think-mode supervision both score \(51.8\), while routing reaches \(53.7\).  For Qwen3-1.7B\(\rightarrow\)0.6B, uniform supervision raises the score from \(15.0\) to \(21.0\), and routing further improves it to \(22.6\).  Teacher headroom alone thus predicts neither the direction nor the magnitude of transfer.  Together with the controlled interventions below, this pattern is consistent with transfer depending on both available benefit and the trace--response interaction.  The Qwen3-1.7B result also raises LCBv5 \bestk{} by \(13.3\) points, broadening solved-problem coverage across code tasks.

\subsubsection{Benefit and Compatibility}
Useful supervision should combine a response that offers outcome benefit with a think trace that yields compatible supervision along that response.  Table~\ref{tab:mechanism_isolation} tests this claim by changing one routing component at a time while fixing the teacher bank, reward-valid responses, normalization, training budget, and checkpoint selection.

The directional interventions separate three explanations.  If arbitrary nonuniform weighting were sufficient, reversing the TRD preference should remain useful; instead, Reversed TRD reaches \(42.5\), below uniform routing at \(44.8\).  If compatibility alone determined utility, removing reward gain should retain the improvement, yet this variant also falls below the uniform control.  Reward-only routing improves over uniform weighting but remains below full \method{} at \(48.0\).  Within this controlled ablation, the observed gain is best explained by combining correction opportunity with the TRD-based compatibility proxy.

The matched controls then replace only the compatibility distance while retaining the reward branch and group normalization.  \emph{Length} uses the number of valid generated tokens and tests whether routing merely favors shorter responses.  \emph{Raw-KL} averages the undiscounted sparse forward KL across the response, testing whether instantaneous disagreement magnitude already captures the effect of TRD.  \emph{Top-\(k\) agreement} averages Teacher--Student support overlap at the scored prefixes and converts it to a distance, testing whether coarse token-set similarity suffices.  All three reuse tensors already produced during distillation and require no additional Teacher or Student forward pass; exact definitions appear in the supplementary material.

TRD attains the highest aggregate result at \(48.0\), exceeding all three matched controls.  Discounted continuation therefore captures routing information beyond response length, instantaneous KL, and coarse token-support agreement in this controlled setting.  The ranking also clarifies the signals' division of labor: reward gain identifies responses with improvement headroom, while TRD differentiates which of them remain coherent with the shared teacher route.

\begin{table}[t]
\centering
\small
\setlength{\tabcolsep}{1.0pt}
\renewcommand{\arraystretch}{1.08}
\begin{tabular*}{\columnwidth}{@{\extracolsep{\fill}}lrrrrr@{}}
\toprule
Configuration & AIME24 & AIME25 & AMC23 & Avg. & \(\Delta\) \\
\midrule
\uniformabbr{} & 29.2 & 25.0 & 57.6 & 44.8 & +0.0 \\
\midrule
\multicolumn{6}{@{}l}{\textit{Routing component ablations}} \\
\quad w/o TRD & 31.0 & 26.3 & 59.4 & 46.5 & +1.7 \\
\quad w/o Reward & 27.5 & 25.4 & 55.6 & 43.4 & -1.4 \\
\quad w/ Reversed TRD & 26.3 & 25.1 & 54.8 & 42.5 & -2.3 \\
\midrule
\multicolumn{6}{@{}l}{\textit{Alternative compatibility proxies}} \\
\quad Raw-KL & 29.8 & 26.2 & 60.3 & 46.8 & +2.0 \\
\quad Length & 32.1 & 25.2 & 59.1 & 46.3 & +1.5 \\
\quad Top-k agreement & 31.9 & 27.7 & 59.7 & 47.2 & +2.4 \\
\midrule
\textbf{\method{} (Ours)} & \textbf{35.2} & \textbf{25.4} & \textbf{60.8} & \textbf{48.0} & \textbf{+3.2} \\
\bottomrule
\end{tabular*}

\caption{Math routing ablations under eval16 (\%, \avgk{}; higher is better).  Avg.\ pools all AIME24, AIME25, and AMC23 problems; \(\Delta\) is relative to \vanillamethod{}.}
\label{tab:mechanism_isolation}
\end{table}

\begin{figure}[t]
\centering
\includegraphics[width=\columnwidth]{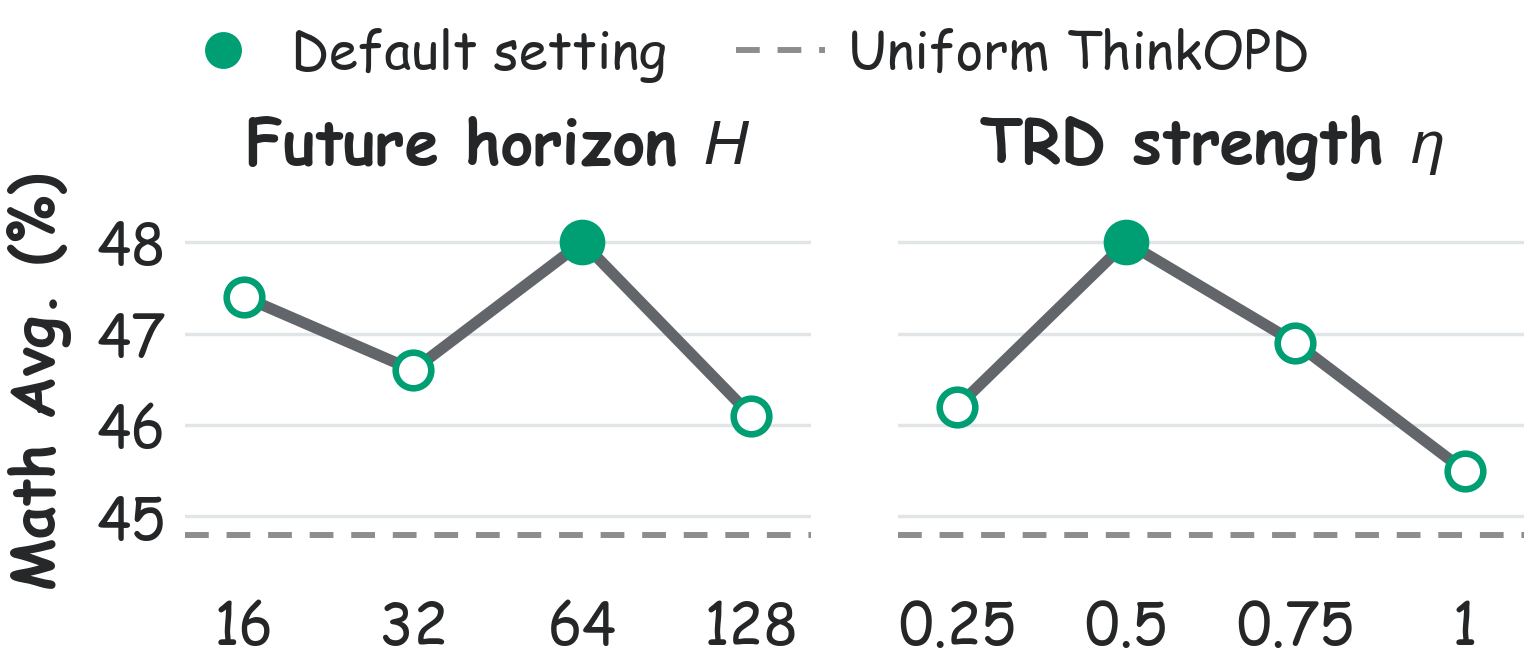}
\caption{Sensitivity of pooled Math \avgk{} to future horizon \(H\) and inverse-TRD strength \(\eta\).  Each sweep varies one parameter while fixing the other at its default.}
\label{fig:param_sensitivity}
\end{figure}

\noindent\textbf{Parameter sensitivity.}
Figure~\ref{fig:param_sensitivity} varies one routing parameter at a time around the default configuration.  Performance stays above uniform routing throughout the tested future horizons, showing robustness over a broad range of continuation windows.  The routing strength \(\eta\) has an interior optimum at \(0.5\): weaker values approach benefit-dominated weighting, while stronger values can suppress responses that combine high TRD with substantial outcome benefit.  This trend reinforces the component analysis by locating the best operating point at a balance between outcome benefit and compatibility.  Across the tested range, the broad tolerance to \(H\) and sharper response to \(\eta\) indicate that temporal aggregation is stable, while balancing benefit against compatibility remains the consequential calibration choice.

\subsubsection{Response Routing Improves Supervision Efficiency}

\begin{figure}[t]
\centering
\includegraphics[width=\columnwidth]{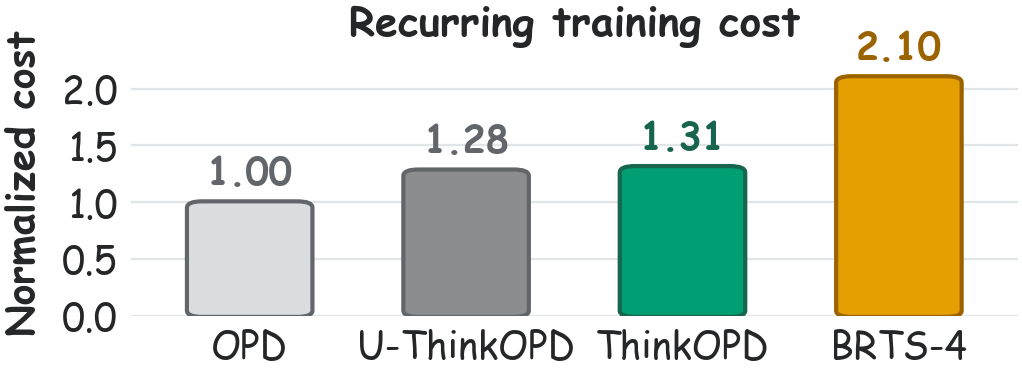}
\caption{Recurring cost under matched asynchronous OPD, normalized to standard OPD (\(1.00\)).  Shared offline-bank construction is excluded; BRTS-4 selects among four online teacher trajectories.}
\label{fig:rq4_cost}
\end{figure}

All methods in Figure~\ref{fig:rq4_cost} use the same asynchronous OPD execution, so the comparison separates three sources of workload rather than different pipeline schedules.  \vanillamethod{} adds the privileged think trace to every scored prefix, lengthening the teacher context and increasing forward and attention/KV processing; its cost is \(1.28\) versus \(1.00\) for standard OPD even though trace generation is offline.  \method{} reaches only \(1.31\) because routing reuses rewards and token-level divergences already computed for distillation, while raising Overall \avgk{} from \(29.0\) to \(32.0\).

BRTS-4 instead generates four candidate teacher trajectories online and selects among them for every group.  It cannot amortize one shared offline trace across sibling responses, increasing recurring cost to \(2.10\).  \method{} reallocates the available trace-conditioned supervision and achieves a \(2.3\)-point higher Overall score with \(37.6\%\) less recurring time.  Detailed accounting appears in the supplement.

\FloatBarrier

\section{Conclusion}

Think-enabled OPD offers a stronger teacher view while preserving student-side exploration, yet its on-policy prefixes do not guarantee that one privileged trace provides equally suitable supervision for every sibling response.  \method{} makes this interaction explicit: reward gain measures outcome benefit, TRD supplies an operational proxy for trace--response compatibility, and group-normalized routing combines them across sibling responses.  Across mathematics and code, this router improves over \vanillamethod{} in both same-model and both cross-model settings and leads the matched external baselines in the representative comparison.  In the controlled ablation, the observed gain is best explained by combining both routing signals, while reusing existing teacher signals keeps the router practical.  \method{} preserves student exploration and allocates the available privileged trace across sampled student paths.  More broadly, think-enabled OPD exposes route-dependent compatibility with a fixed teacher, providing a controlled setting for studying supervision transfer across student paths.

\bibliography{tnt_opd_refs}

\end{document}